\documentclass[runningheads]{llncs}
\usepackage{graphicx}
\usepackage{times}
\usepackage{epsfig}
\usepackage{amsmath}
\usepackage{amssymb}
\usepackage{booktabs}
\usepackage{multirow}
\usepackage{graphicx}
\usepackage{xcolor}

\begin{document}
\title{ContraReg: Contrastive Learning of Multi-modality Unsupervised Deformable Image Registration}
\titlerunning{ContraReg: Contrastive Multi-modality Registration}
%
\makeatletter
\newcommand{\printfnsymbol}[1]{%
  \textsuperscript{\@fnsymbol{#1}}%
}
\makeatother

\author{Neel Dey\index{Dey, Neel}\inst{1} \and
Jo Schlemper\index{Schlemper, Jo}\inst{2} \and
Seyed Sadegh Mohseni Salehi\index{Sadegh Mohseni Salehi, Seyed}\inst{2} \and
Bo Zhou\index{Zhou, Bo}\inst{3} \and
Guido Gerig\index{Gerig, Guido}\inst{1} \and
Michal Sofka\index{Sofka, Michal}\inst{2}}

\authorrunning{N. Dey, et al.}

\institute{
Department of Computer Science \& Engineering, New York University, Brooklyn, NY, USA. \and
Hyperfine Research, Guilford, CT, USA. \and
Department of Biomedical Engineering, Yale University, New Haven, CT, USA.}

%
%
\maketitle
\begin{abstract}

Establishing voxelwise semantic correspondence across distinct imaging modalities is a foundational yet formidable computer vision task. Current multi-modality registration techniques maximize hand-crafted inter-domain similarity functions, are limited in modeling nonlinear intensity-relationships and deformations, and may require significant re-engineering or underperform on new tasks, datasets, and domain pairs.
This work presents ContraReg, an unsupervised contrastive representation learning approach to multi-modality deformable registration. By projecting learned multi-scale local patch features onto a jointly learned inter-domain embedding space, ContraReg obtains representations useful for non-rigid multi-modality alignment. Experimentally, ContraReg achieves accurate and robust results with smooth and invertible deformations across a series of baselines and ablations on a neonatal T1-T2 brain MRI registration task with all methods validated over a wide range of deformation regularization strengths.

\end{abstract}

\section{Introduction}

The spatial alignment (or \textit{registration}) of images from sources capturing distinct anatomical characteristics enables well-informed biomedical decision-making via multi-modality information integration. For example, multi-modality registration of intra-operative to pre-operative imaging is crucial to various surgical procedures~\cite{hata1998multimodality,nimsky2006intraoperative,risholm2011multimodal}. Consequently, several inter-domain image similarity functions have been developed to drive iterative or learning-based registration~\cite{haber2006intensity,heinrich2012mind,wells1996multi}.
Yet, despite the decades-long development of multi-modality objectives, accurate deformable registration of images with highly nonlinear relationships between their appearance and shapes remains difficult. 

Losses operating on intensity features (global and local 1D histograms~\cite{guo2019multi,russakoff2004image,studholme1999overlap,wells1996multi}, local descriptors~\cite{heinrich2012mind,wachinger2012entropy,woo2014multimodal}, edge-maps~\cite{haber2006intensity}, among others) are typically hand-crafted and do not consistently generalize outside of the domain-pair they were originally proposed for and necessitate non-trivial domain expertise to tune towards optimal results. 
More recent multi-modality methods based on \textit{learned} appearance similarity~\cite{pielawski2020comir}, simulation-driven semantic similarity~\cite{hoffmann2021synthmorph}, and image translation~\cite{qin2019unsupervised} have demonstrated strong registration performance but may only apply to supervised affine registration~\cite{pielawski2020comir}, require population-specific segmentation labels for optimal registration~\cite{hoffmann2021synthmorph}, or necessitate extensive and delicate GAN-based training frameworks~\cite{qin2019unsupervised}.

To overcome these limitations, this work develops \verb|ContraReg| (CR), an unsupervised representation learning framework for non-rigid multi-modality registration. CR requires that, once registered, corresponding positions and patches in the moved and fixed images have \textit{high mutual information} in a jointly-learned multi-scale multi-modality embedding space. These characteristics are achieved via self-supervised contrastive training and only requires an unsupervised feature extractor (which can be pretrained or randomly-initialized and frozen). As a result, on the challenging task of neonatal inter-subject T1w-T2w diffeomorphic MRI registration, CR achieves higher anatomical overlap with comparable deformation characteristics to previous similarity metrics, validated over a wide range of regularization parameters. Finally, for experimental completeness, we then evaluate CR across a diversity of auxiliary losses and external negative sample selection and pretraining strategies.

\begin{figure}[t]
    \centering
    \includegraphics[width=\textwidth]{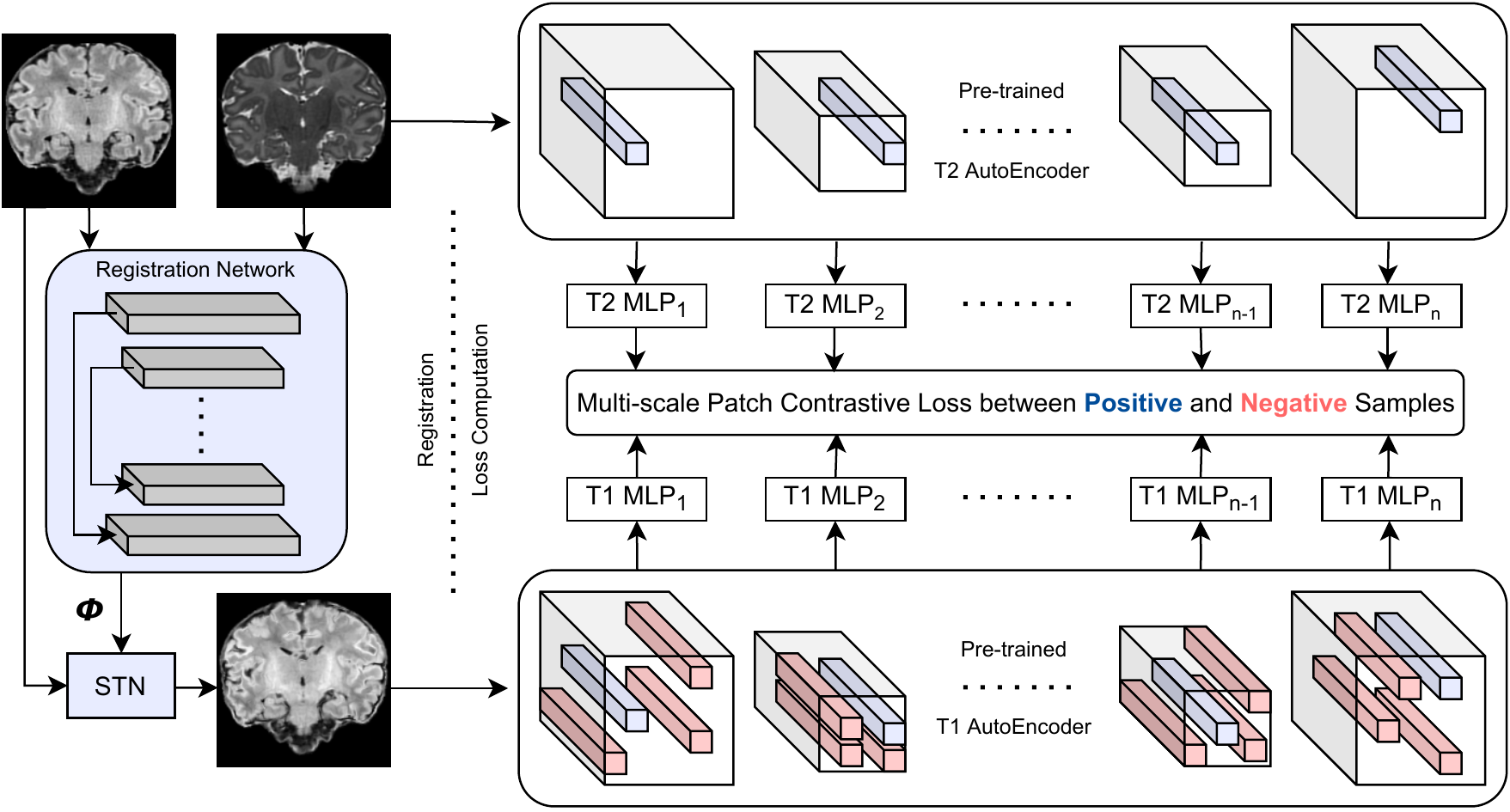}
    \caption{\textbf{ContraReg}. Given displacements and a moved image from a multi-modality registration network (\textbf{left}), a contrastive loss is calculated on multi-scale patches extracted from modality-specific networks (\textbf{right}), such that corresponding locations have high mutual information. In practice, our implementation is bidirectional s.t. the inverse modality-pair loss is also computed.}
    \label{fig:overview}
\end{figure}

\section{Related work}

\noindent\textbf{Hand-crafted similarity losses}. Mutual information and its variants~\cite{loeckx2009nonrigid,studholme1999overlap,wells1996multi} typically operate on image intensity histograms and may be limited in their ability to model complex non-rigid deformations. 
Conversely, local mutual information methods~\cite{guo2019multi,russakoff2004image} are spatially-aware but may not have enough samples to build accurate patch intensity histograms.
Other losses operating on intensity-derived local  descriptors~\cite{heinrich2012mind,woo2014multimodal} or edges~\cite{haber2006intensity,wachinger2012entropy} learn domain-invariant features and have been successful in tasks such as body cavity MR-CT registration with relatively limited adoption in neuroimaging. \\

\noindent\textbf{Simulation}. Registration via translation approaches seek to simulate one modality from the other, such that the problem can be reduced to a mono-modality alignment~\cite{arar2020unsupervised,qin2019unsupervised,zhou2021anatomy}. While performant, recent methods can be susceptible to the hallucinatory or instability drawbacks of medical image translation leading to suboptimal alignment~\cite{lu2021image,ren2021harmonization}. 
Recently, SynthMorph~\cite{hoffmann2021synthmorph} simulates both deformations and synthetic appearances for neuroimages to train a general-purpose multi-modality  network. \verb|ContraReg| instead obtains highly accurate warps for a given dataset at the cost of dataset-specific training. \\

\noindent\textbf{Learned similarity losses}. Finally, several methods attempt to learn an inter-domain similarity metric via supervised learning with pre-aligned training images~\cite{gutierrez2017guiding,lee2009learning,pielawski2020comir,simonovsky2016deep}. For example, CoMIR~\cite{pielawski2020comir} uses supervised contrastive learning to learn affine registration between highly visually disparate modalities. 
In particular, \verb|ContraReg| draws inspiration from PatchNCE~\cite{park2020contrastive}, an image translation framework using contrastive losses on multi-scale patches. However, a straightforward extension of PatchNCE to registration is not possible and would lead to degenerate identity solutions to the PatchNCE loss as we require a warp between two distinct input images~\cite{jing2021understanding}. 
This work presents a different approach to incorporating multi-scale patches via \textit{externally-trained} feature extractors which enables successful registration without degenerate solutions.

\section{Methods}

Fig. \ref{fig:overview} illustrates \verb|ContraReg|. Architectural details are in the supplementary material. \\

\noindent\textbf{Unsupervised pre-training.} To extract multi-scale $n$-dimensional features in an unsupervised manner, we first train modality-specific autoencoders $A_{1}$ and $A_{2}$ as domain-specific features can be beneficial to patchwise contrastive training~\cite{han2021dualcontrastive}. Training is done on $128^{3}$ crops with random flipping and brightness/contrast/saturation augmentation, using an $\mathcal{L}_1$ + Local NCC (window width $ = 7$ voxels) reconstruction loss~\cite{avants2011reproducible}. \\

\noindent\textbf{Registration training.} This work focuses on unsupervised learning of registration and multi-modality similarity. Given cross-modality volumes $I_{1}$ and $I_{2}$, a VoxelMorph-style~\cite{dalca2019diffeo} U-Net with constant channel width ($ch$) predicts a stationary velocity field $v$, which when numerically integrated with $ts$ time steps, yields an approximately diffeomorphic displacement $\phi$. We focus on bidirectional registration and obtain the inverse warp via integrating $-v$. This network is trained with a $(1 - \lambda)\mathcal{L}_{sim} + \lambda \mathcal{L}_{reg}$ objective where $\mathcal{L}_{reg} = \|v\|_2^2$ is a regularizer controlling velocity (and indirectly displacement) field smoothness, $\mathcal{L}_{sim} = \frac{1}{2} (d_{12}(I_{1} || I_{2} \circ \phi) + d_{21}(I_{2} || I_{1} \circ \phi^{-1}))$ is a registration term s.t. $d_{12, \ 21}$ measures inter-domain similarity, and $\lambda$ is a hyperparameter.

Here, we define $d_{12}$ and $d_{21}$ is defined analogously. We first extract multi-scale spatial features $A_{1}^{k}(I_{1})$ and $A_{2}^{k}(I_{2} \circ \phi)$ using the autoencoders, where $k = 1, ..., L$ is the layer index and $L$ is the number of layers. 
A perceptual registration loss~\cite{czolbe2021semantic} is inappropriate in this setting as these features correspond to different modality-specific spaces. 
Instead, we maximize a lower bound on mutual information between corresponding spatial locations in $A_{1}^{k}(I_{1})$ and $A_{2}^{k}(I_{2} \circ \phi)$ by minimizing a noise contrastive estimation loss~\cite{oord2018representation}. As in~\cite{park2020contrastive}, we project the channel-wise autoencoder features of size $\mathbb{R}^{N^k \times C^k}$ (where $N^k$ is the number of spatial indices and $C^k$ is the number of channels in layer $k$) onto a hyperspherical representation space to obtain features $F_{1}^{k}(A_{1}^{k}(I_{1}))$ and $F_{2}^{k}(A_{2}^{k}(I_{2} \circ \phi))$ where $F_{1, 2}$ are 3-layer 256-wide trainable ReLU-MLPs~\cite{chen2020big}. In this space, indices in correspondence $f_{i}^{k} = F_{1}^{k}((A_{1}^{k}(I_{1}))_i)$ and $f_{i}^{k+} = F_{2}^{k}((A_{2}^{k}(I_{2} \circ \phi))_i)$, where $i = 1, ..., N^{k}$, are positive pairs. Similarly, $f_{i}^{k}$ and $f_{j}^{k-} = F_{2}^{k}((A_{2}^{k}(I_{2} \circ \phi))_j)$, where $j = 1, ..., N^{k}$ and $j \neq i$, are negative pairs. 

For optimal contrastive learning, we sample a single positive pair and $ns >> 1$ negative samples, and use the following loss with $\tau$ as a temperature hyperparameter:
\begin{equation*}
    d_{12}(I_{1} || I_{2} \circ \phi) = \sum_{k=1}^{L} \sum_{i=1}^{N^{k}} - \text{log} \Bigg( \frac{\text{exp}(f_{i}^{k} \cdot f_{i}^{k+} / \tau)}{\text{exp}(f_{i}^{k} \cdot f_{i}^{k+} / \tau) + \sum_{j=1, j \neq i}^{ns} \text{exp}(f_{i}^{k} \cdot f_{j}^{k-} / \tau)} \Bigg)
\end{equation*}
Notably, as medical images also acquire empty space outside of the body, random patch sampling will lead to the sampling of false positive and negative pairs (e.g., background voxels sampled as both positive and negative pairs). To this end, the masked CR (\verb|mCR|) model samples voxel pairs only within the union of the binary foregrounds of $I_{1}$ and $I_{2}$ and resizes this mask to the layer-$k$ specific resolution when sampling from $A_{1}^{k}(I_{1})$ and $A_{2}^{k}(I_{2} \circ \phi)$. We further investigate tolerance to false positive and negative training pairs and thus also train models without masking (denoted \verb|CR| only).  \\

\noindent\textbf{Hypernetwork optimization.} Registration performance strongly depends on weighing $\lambda$ correctly for a given dataset and $\mathcal{L}_{sim}$. Therefore, for fair comparison, the entire range of $\lambda$ is evaluated for all benchmarked methods using hypernetworks~\cite{ha2016hypernetworks} developed for registration~\cite{hoopes2021hypermorph,mok2021conditional}. Specifically, the FiLM~\cite{perez2018film} based framework of \cite{mok2021conditional} is used with a 4-layer 128-wide ReLU-MLP to generate a $\lambda \sim \mathcal{U}[0, 1]$-conditioned shared embedding, which is then linearly projected (with a weight decay of $10^{-5}$) to each layer in the registration network to generate $\lambda$-conditioned scales and shifts for the network activations. At test time, we sample 17 registration networks for each method with dense $\lambda$ sampling between $[0, 0.2)$ and sparse sampling between $[0.2, 1.0]$. 

\begin{figure}[!ht]
    \centering
    \includegraphics[width=\textwidth]{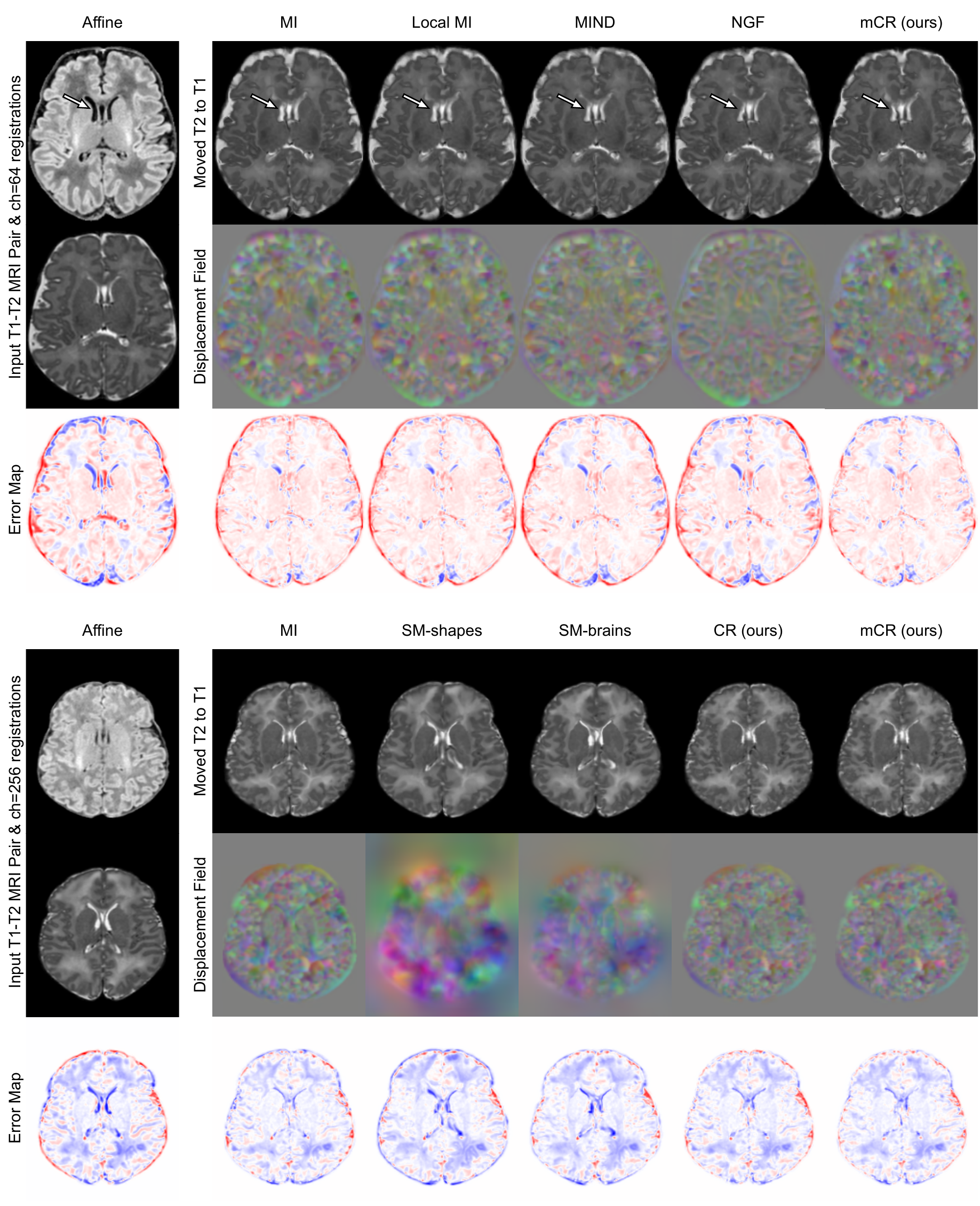}
    \caption{T1w-T2w registration visualization between arbitrarily selected subjects for the (\textbf{top}) ch=64 and (\textbf{bottom}) ch=256 models. Error maps computed w.r.t. the T2w MRI of the target subject. Hypernetwork registration models are sampled with the same $\lambda$ as Table \ref{table:main_quant}.}
    \label{fig:results_qual}
\end{figure}

\begin{figure}[t]
    \centering
    \includegraphics[width=\textwidth]{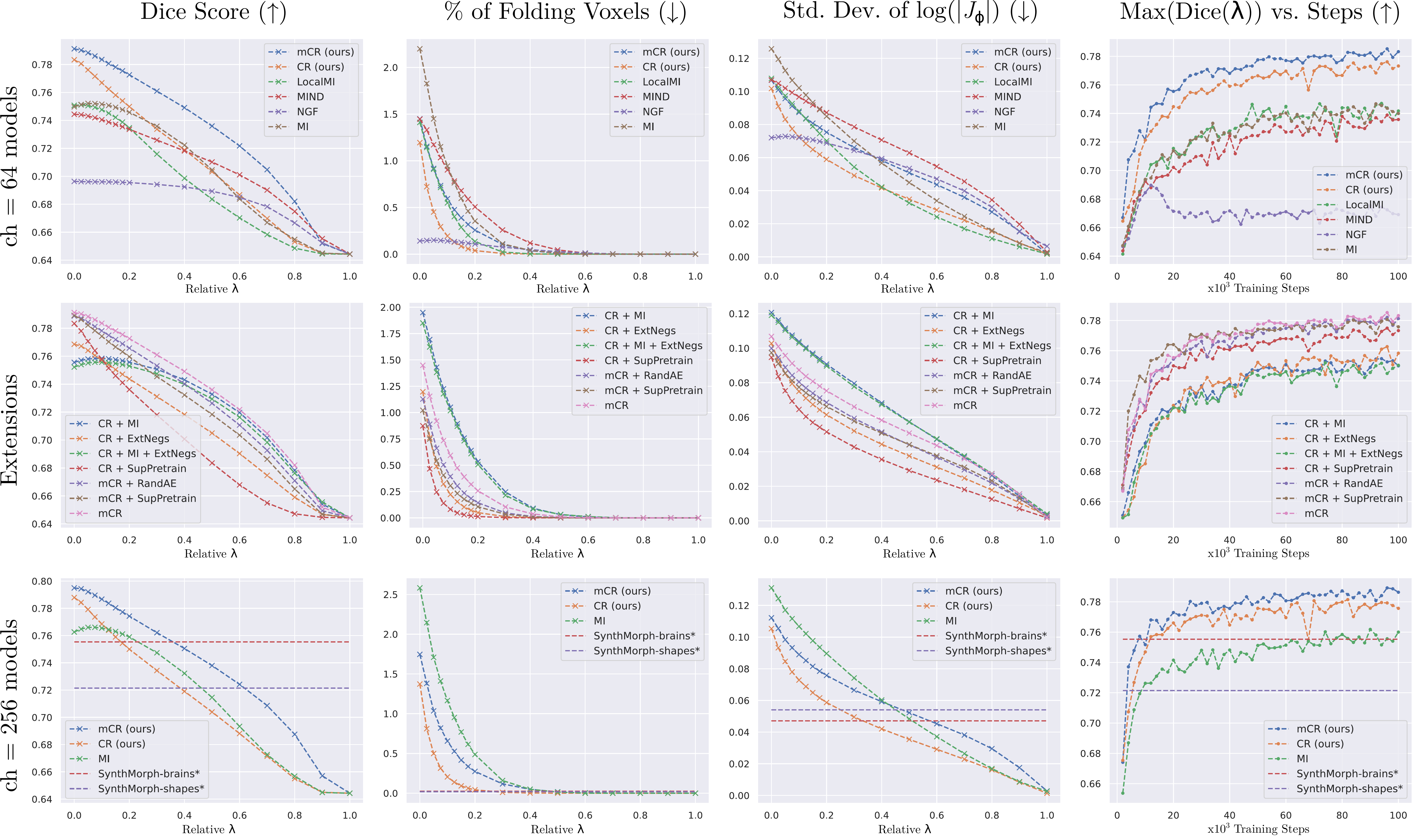}
    \caption{Plots of registration accuracy vs. $\lambda$ (\textbf{col. 1}), deformation qualities vs. $\lambda$ (\textbf{cols. 2 \& 3}), and accuracy vs. training steps (\textbf{col. 4}). Benchmarks are performed against commonly used multi-modality losses (\textbf{row 1}), extensions of the proposed techniques (\textbf{row 2}), and recent modality-pair agnostic methods (\textbf{row 3}). Across baseline losses, CR and mCR achieve the best tradeoff between accuracy and deformation characteristics (\textbf{row 1, cols. 1--3}). Further, using external losses and/or negatives reduces performance and supervised pretraining does not yield notable improvements (\textbf{row 2, cols. 1--3}). Compared to SynthMorph-brains~\cite{hoffmann2021synthmorph}, CR and mCR obtain higher accuracy (\textbf{row 3, col. 1}) in the $\lambda = 0.0 - 0.15$ and $0.0 - 0.3$ ranges, respectively, at the cost of more irregular warps (\textbf{row 3, cols. 2--3}). See Table \ref{table:main_quant} for an analysis of trading off accuracy for smoothness.}
    \label{fig:plot_quant}
\end{figure}

\section{Experiments}

\noindent\textbf{Data.} We compare registration methods by benchmarking on inter-subject multi-modality registration. We use pre-processed T1w and T2w MRI of newborns imaged at 29-45 weeks gestational age from  dHCP~\cite{makropoulos2018developing}, a challenging dataset due to rapid temporal development in morphology and appearance alongside inter-subject variability. We follow~\cite{dey2021templates} for further preprocessing to obtain $160 \times 192 \times 160$ volumes at $0.6132 \times 0.6257 \times 0.6572 \ mm^3$ resolution and use 405/23/64 images for training, validation, and testing. \\

\noindent\textbf{Evaluation methods.} Registration evaluation is non-trivial due to competing objectives. Low smoothness regularization ($\lambda$) can allow for near-exact matching of appearance but with anatomically-implausible and highly irregular deformations. Conversely, high $\lambda$ enables smooth deformations with suboptimal alignment. Therefore, we evaluate registration performance and robustness as a function of $\lambda$, via Dice and Dice30 (average of 30\% of lowest dice scores) scores, respectively, calculated between the target and moved label maps of the input images (segmentations provided by dHCP~\cite{makropoulos2014automatic}). To investigate deformation smoothness, we also evaluate the percentage of folding voxels and the standard deviation of the log Jacobian determinant of $\phi$~\cite{hering2021learn2reg} as a function of $\lambda$. \\

\noindent\textbf{Benchmarked methods.} Using the same registration network with $ch = 64$ and $ts = 5$, we benchmark popular multi-modality metrics including Mutual Information (MI) (48 bins), local MI (48 bins, patch size = 9), MIND (distance = 2, patch size = 3), and normalized gradient fields, alongside the proposed \verb|mCR| and \verb|CR| models. We further compare against the general-purpose \verb|SynthMorph| (SM)-shapes and brains models~\cite{hoffmann2021synthmorph}, by using their publicly released models and affine-aligning the images to their atlas. As SM uses $ch = 256$, we retrain the proposed registration models at that width. As we use public models, we cannot perform $\lambda$-conditioning and evaluation for SM. Further, as inter-subject dHCP registration can require large non-smooth deformations, we study whether a higher number of integration steps improves deformation characteristics (as in \cite{schuh2018thesis}) for the $ch=256$ model, evaluating $ts = \{10, 16, 32\}$ with $32$ as default. 

To evaluate extensions of \verb|CR| and \verb|mCR|, we investigate whether combining them with a global loss (\textit{+ MI}), incorporating more negative samples from an external randomly-selected subject (\textit{+ ExtNegs}), or both (\textit{+ MI + ExtNegs}) lead to improved results. We then evaluate the importance of feature extractor pretraining by using randomly-intialized \textit{frozen} autoencoders instead as a worst-case feature extractor (\textit{+ RandAE}). Finally, we evaluate whether contrastively pre-training the autoencoders and projection MLPs by using ground truth multi-modality image pairs alongside the reconstruction losses (\textit{+ SupPretrain}) leads to improved results, with the following loss, where $I_{1, 2}$ are from the same subject, $\lambda_{sp} = 0.1$, and $\hat{I_{1,2}}$ are the reconstructions,
\begin{equation*}
    \mathcal{L}_{A_{1}, A_{2}, F_{1}, F_{2}} = \lambda_{sp} d_{12}(I_{1}, I_{2}) + \sum_{i=1}^2 (\|I_{Ti} - \hat{I_{Ti}}\|_1 + NCC(I_{Ti}, \hat{I_{Ti}})).
\end{equation*}

\noindent\textbf{Implementation details.} All models were developed in TensorFlow 2.4 and were all trained for $10^5$ iterations with Adam on a V100 GPU. For stability across all methods, we use a conservative learning rate of $5 \times 10^{-5}$. For the contrastive loss, we set $ns = 1024$ and $\tau = 0.007$. The autoencoder has 7 \verb|Conv-IN-LeakyReLU(0.2)| blocks with 3 down/up sampling layers and 32-64-128-64-32-32-32-1 filters with its post-convolution features from the first 6 layers sampled for the contrastive loss.  In practice, to avoid tuning the sampling strategy for $\lambda$ as in \cite{hoopes2021hypermorph}, we add a rescaling constant $\alpha = 0.1$ to the objective function for \verb|CR| and \verb|mCR| with the form $\alpha (1 - \lambda)\mathcal{L}_{sim} + \lambda \mathcal{L}_{reg}$. \\

\begin{table*}[!t]
\caption{\textbf{Trading off performance for invertibility}. Registration accuracy (Dice), robustness (Dice30), and characteristics (\% Folds, stddev. log$|J_{\varphi}|$) for all benchmarked methods at values of $\lambda$ that maintains the percentage of folding voxels at less than $0.5\%$ of all voxels, as in~\cite{qiu2021learning}, s.t. high performance is achieved alongside negligible singularities. This table is best interpreted in conjunction with figure \ref{fig:plot_quant}, where results from all $\lambda$ values are visualized. \textbf{A.} CR and mCR obtain improved accuracy and robustness (\textit{A5-6}) with similar deformation characteristics to baseline losses (\textit{A1-4}). \textbf{B.} At larger model sizes, mCR and CR still obtain higher registration accuracy and robustness (\textit{B4-5}), albeit at the cost of more irregular deformations in comparison to SM (\textit{B1}). \textbf{C.} Further adding external losses, negative samples, or both to CR harms performance (\textit{C1-3}), supervised pretraining (\textit{C4-5}) very marginally improves results over training from scratch (\textit{A5-6}), and random feature extraction only slightly reduces Dice while smoothening displacements (\textit{C6}). \textbf{D.} At a given $\lambda$, increasing integration steps yields marginal Dice and smoothness improvements.}
\centering
\begin{tabular}{cclcccccc}
\toprule
\textbf{Set} & \textbf{Width} & \textbf{Method} & \textbf{Opt.} $\lambda$ & \textbf{Dice} ($\uparrow$) & \textbf{Dice30} ($\uparrow$) & \textbf{\textbf{\% Folds}} ($\downarrow$) & \textbf{sdlog}$|J_{\varphi}|$ ($\downarrow$) \\ \midrule
A & 64 & NGF~\cite{haber2006intensity} & $0.0$ & $0.696 \pm 0.023$ & $0.686$ & $\mathbf{0.141} \pm 0.043$ & $\mathbf{0.072}$ \\
& 64 & MI~\cite{wells1996multi} &  $0.175$ & $0.748 \pm 0.021$ & $0.739$ & $0.461 \pm 0.100$ & $0.089$  \\
& 64 & LocalMI~\cite{guo2019multi} & $0.125$ & $0.745 \pm 0.023$ & $0.737$ & $0.402 \pm 0.076$ & $0.083$ \\
& 64 & MIND~\cite{heinrich2012mind} & $0.3$ & $0.726 \pm 0.023$ & $0.716$ & $0.258 \pm 0.051$ & $0.079$ \\
& 64 & CR (proposed) & $0.05$ & $0.776 \pm 0.020$ & $0.768$ & $0.451 \pm 0.074$ & $0.083$ \\
& 64 & mCR (proposed) & $0.125$ & $\mathbf{0.781} \pm 0.020$ & $\mathbf{0.774}$ & $0.475 \pm 0.070$ & $0.084$ \\ \midrule

B & 256 & SM-brains~\cite{hoffmann2021synthmorph} & - & $0.755 \pm 0.020$ & $0.749$ & $0.023 \pm 0.008$ & $\mathbf{0.048}$ \\
& 256 & SM-shapes~\cite{hoffmann2021synthmorph} & - & $0.721 \pm 0.021$ & $0.715$ & $\mathbf{0.017} \pm 0.011$ & $0.056$ \\
& 256 & MI~\cite{wells1996multi} & $0.2$ & $0.759 \pm 0.021$ & $0.750$ & $0.487 \pm 0.099$ & $0.090$ \\
& 256 & CR (proposed) & $0.075$ & $0.774 \pm 0.020$ & $0.765$ & $0.315 \pm 0.0576$ & $0.078$ \\
& 256 & mCR (proposed) & $0.15$ & $\mathbf{0.780} \pm 0.021$ & $\mathbf{0.773}$ & $0.416 \pm 0.065$ & $0.082$ \\
\midrule
C & 64 & CR+MI & $0.3$ & $0.751 \pm 0.021$ & $0.742$ & $0.246 \pm 0.059$ & $0.080$ \\
& 64 & CR+ExtNegs & $0.05$ & $0.764 \pm 0.020$ & $0.756$ & $0.489 \pm 0.073$ & $0.085$\\
& 64 & CR+MI+ExtNegs & $0.3$ & $0.747 \pm 0.021$ & $0.739$ & $0.214 \pm 0.056$ & $0.078$\\
& 64 & CR+SupPretrain & $0.025$ & $0.778 \pm 0.020$ & $0.770$ & $0.465 \pm 0.075$ & $0.084$ \\
& 64 & mCR+SupPretrain & $0.075$ & $0.778 \pm 0.020$ & $0.770$ & $0.406 \pm 0.067$ & $0.081$ \\
& 64 & mCR+RandAE & $0.1$ & $0.778 \pm 0.020$ & $0.770$ & $0.393 \pm 0.070$ & $0.80$ \\ \midrule
D & 256 & CR (10 int. steps) & $0.075$ & $0.773 \pm 0.021$ & $0.764$ & $0.341 \pm 0.058$ & $0.079$ \\
& 256 & CR (16 int. steps) & $0.05$ & $0.779 \pm 0.020$ & $0.772$ & $0.462 \pm 0.071$ & $0.083$ \\
& 256 & CR (32 int. steps) & $0.075$ & $0.774 \pm 0.020$ & $0.765$ & $0.315 \pm 0.0576$ & $0.078$ \\
\bottomrule

\end{tabular}
\label{table:main_quant}
\end{table*}

\noindent\textbf{Results.} Sample registration visualizations are provided in Fig. \ref{fig:results_qual}, performance scores versus $\lambda$ are plotted in Fig. \ref{fig:plot_quant}, and a study of trading-off registration accuracy for smoothness is tabulated in Table \ref{table:main_quant}. We make the following experimental observations:

\begin{itemize}
    \item \textit{(m)CR achieves higher accuracy and converges faster than baseline losses}. Fig. \ref{fig:plot_quant} (row 1) indicates that the proposed models achieve better Dice with comparable (mCR) or better (CR) folding and smoothness characteristics in comparison to baseline losses as a function of the 17 values of $\lambda$ tested.  Further, Table \ref{table:main_quant} reveals that if anatomical overlap is reduced to also achieve negligible folding (defined as folds in 0.5\% of all voxels~\cite{qiu2021learning}), CR and mCR still achieve the optimal tradeoff.

    \item \textit{(m)CR achieves more accurate registration than label-trained methods at the cost of lower warp regularity}. While the public SM-brains model does not achieve the same Dice score as (m)CR, it achieves the third-highest performance behind (m)CR with substantially smoother deformations. We postulate that this effect stems from the intensity-invariant label-based training of SM-brains only looking at the semantics of the image, whereas our approach and other baselines are appearance based.

    \item \textit{Masking consistently improves results}. Excluding false positive and false negative pairs from the training patches yields improved registration performance across all values of $\lambda$ with acceptable increases in deformation irregularities vs. $\lambda$ (Fig \ref{fig:plot_quant} rows 1 \& 3; cols 1-3). Importantly, contrastive training without foreground masks (CR) still outperforms other baseline losses and does so with smoother warps.
    
    \item \textit{Pretraining has a marginal impact on (m)CR.} While still outperforming other baselines, using a \textit{randomly-intialized \& frozen}  feature extractor achieves marginally lower dice with longer convergence times as compared to the full pretrained model.
    
    \item \textit{Using external losses or negatives with (m)CR does not improve results}. Combining a global loss (MI) with CR does not improve results, which we speculate is due to the inputs already being globally affine-aligned. We also see a similar phenomenon to~\cite{park2020contrastive}, where adding external negatives from other subjects lowers performance.
    
    \item \textit{Self-supervision yields nearly the same performance as supervised pretraining}. Comparing rows A5-6 and C4-5 of Table \ref{table:main_quant} reveals that utilizing supervised pairs of aligned images for pretraining $A_{1,2}$ and $F_{1,2}$ yields very similar results, indicating that supervision is not required for optimal registration in this context.

\end{itemize}

\section{Discussion}
\noindent\textbf{Limitations and future work.} Some limitations exist in the presented material and will be addressed in subsequent work: 
(1) While we perform $\lambda$-conditioned hypernetwork registration to fairly compare benchmarked losses across all regularization strengths, hypernetworks may not exactly approximate all $\lambda$ conditions. Further, hypernetworks were not trained for two of our baselines (SM-brains and shapes~\cite{hoffmann2021synthmorph}) as we instead used their public models and we regularize for \textit{velocity}-smoothness instead of \textit{warp}-smoothness as in their work, both of which confound comparisons.
(2) It is probable that combining our appearance-based approach with label-based simulation~\cite{hoffmann2021synthmorph} would further improve results.  
(3) We did not explore other architectural configurations for the autoencoders and MLPs and it is plausible that there may be significant room for optimization. (4) We benchmarked the simulated inter-subject registration task and other use-cases such as pre-to-intra operative warping and preprocessing for multi-sensor fusion~\cite{dey2019multi} may show different trends. (5) (m)CR currently requires $\sim15\%$ more time per training iteration w.r.t. mutual information and can be optimized. (6) Unsupervised patch sampling may introduce false negative pairs in the contrastive loss and can be avoided with unsupervised \textit{negative-free} patch representation learning methods~\cite{ren2022local}. \\

\noindent\textbf{Conclusions.} This work presented \verb|ContraReg|, a self-supervised contrastive representation learning approach to diffeomorphic non-rigid image registration. On the challenging task of inter-subject T1w-T2w registration with neonatal images showing high appearance and morphological variation, CR achieved high registration performance and robustness while maintaining desirable deformation qualities such as invertibility and smoothness. Finally, CR was validated across several baseline losses (including MI, MIND, NGF), training configurations, and frameworks, with results indicating that training supervision, losses, or external negative sampling strategies are not required. \\

\noindent\textbf{Acknowledgements.} N. Dey and G. Gerig acknowledge partial support from NIH 1R01HD088125‐01A1. Quantitative benchmarking was performed using data made available from the Developing Human Connectome Project which is funded by the European Research Council under the European Union’s Seventh Framework Programme (FP/2007-2013) / ERC Grant Agreement no. [319456]”.

\bibliographystyle{splncs04}
\bibliography{egbib}

\clearpage
\section*{Supplementary Material}

\begin{figure}[!h]
    \centering
    \includegraphics[width=\textwidth]{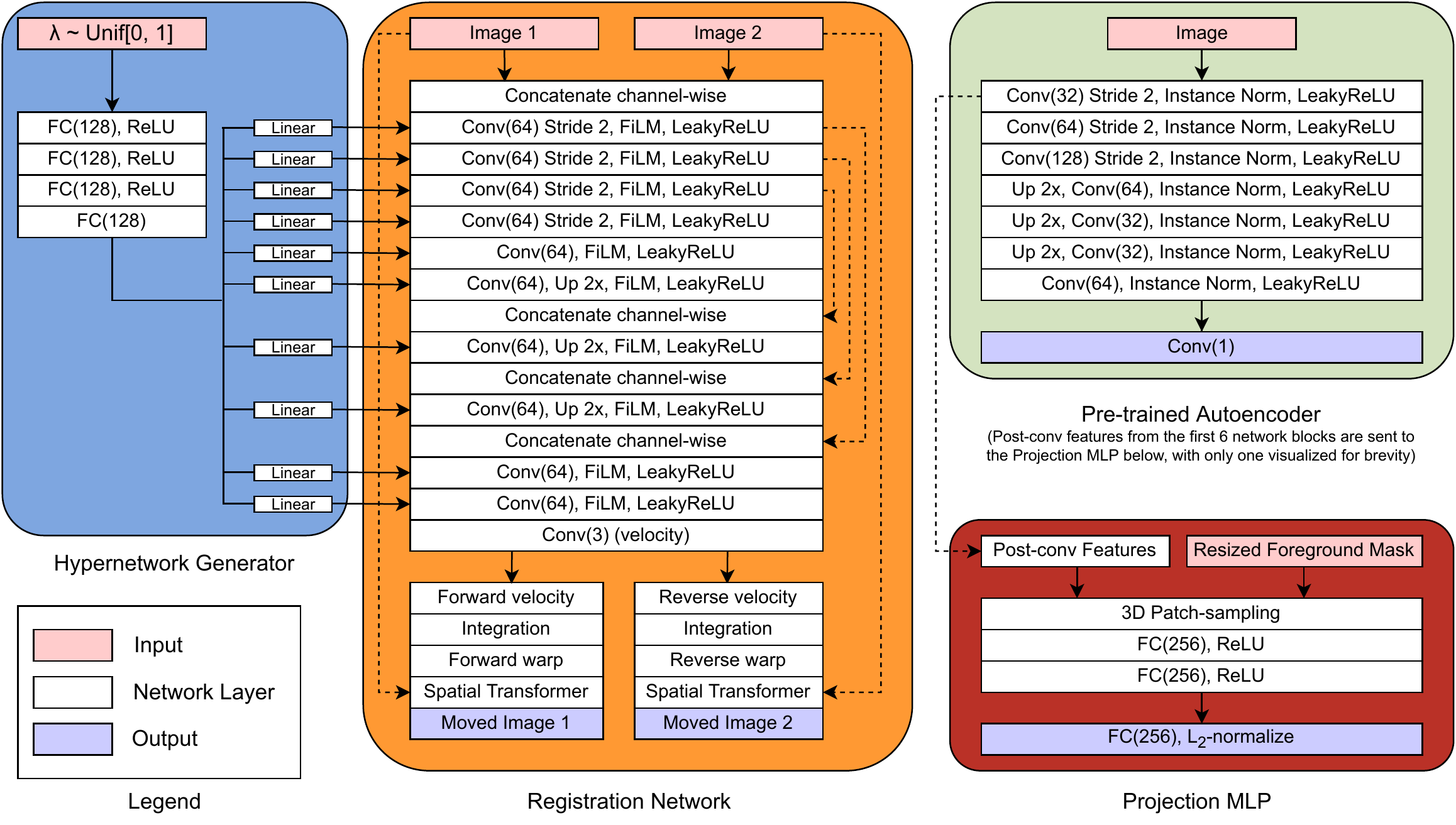}
    \caption{\textbf{Low-level architectures} used in this paper. Connections between the registration network, autoencoder, and projection MLPs correspond to Figure 1 in the main text. Every convolution uses a 3x3x3 kernel size. LeakyReLU slopes set to 0.2.}
    \label{fig:archs}
\end{figure}

\vspace{-2em}

\begin{figure}[!hb]
    \centering
    \includegraphics[width=\textwidth]{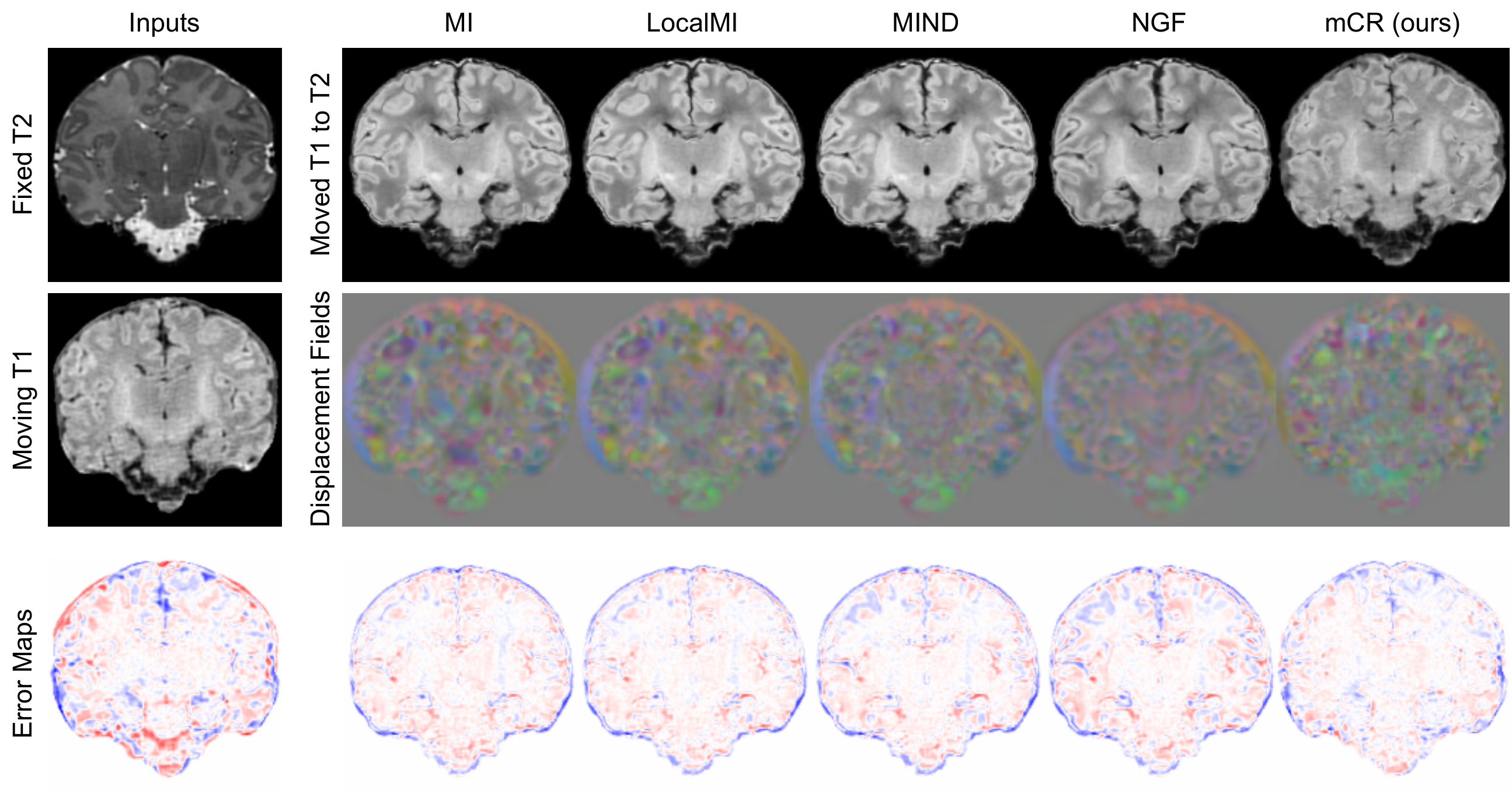}
    \caption{Coronal view of \textbf{T1w-T2w registration} between arbitrarily selected subjects for the ch=64 models. Error maps computed w.r.t. the T1w MRI of the target subject. Hypernetwork registration models are sampled with the same $\lambda$ as Table 1 (main text).}
    \label{fig:coronalcollage}
\end{figure}

\begin{figure}[t]
    \centering
    \includegraphics[width=\textwidth]{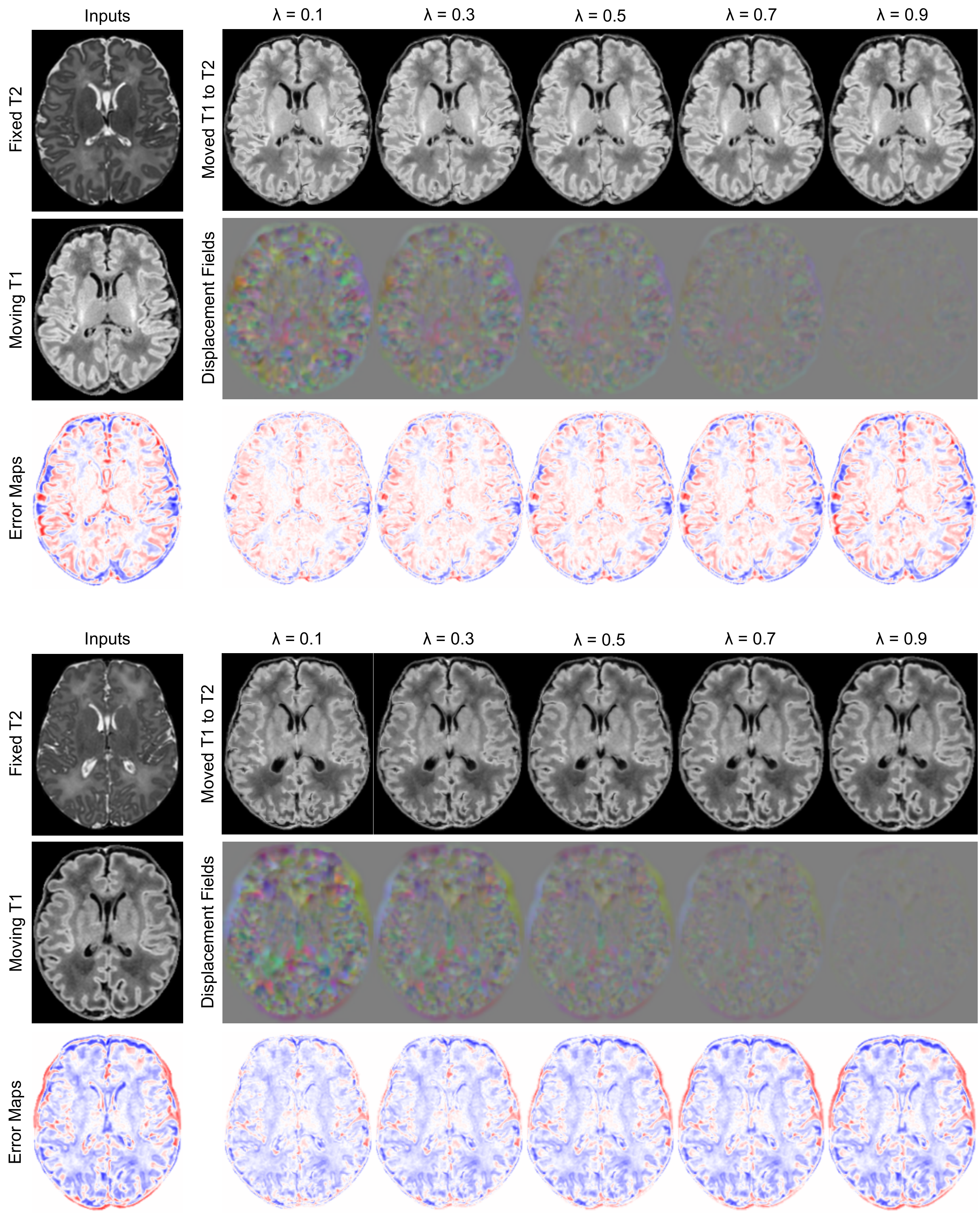}
    \caption{\textbf{Hypernetwork T1w-T2w registration}. \textbf{Leftmost column:} Arbitrarily selected input images to register. \textbf{Remaining columns:} Registration results in ascending order of regularization strengths ($\lambda$). As $\lambda$ increases, deformation energies reduce and regularity increases, at the cost of increased image mismatch.}
    \label{fig:hypernet}
\end{figure}

\end{document}